\documentclass{article}

\usepackage{arxiv}
\usepackage[utf8]{inputenc} % allow utf-8 input
\usepackage[T1]{fontenc}    % use 8-bit T1 fonts
\usepackage{hyperref}       % hyperlinks
\usepackage{url}            % simple URL typesetting
\usepackage{booktabs}       % professional-quality tables
\usepackage{amsfonts}       % blackboard math symbols
\usepackage{nicefrac}       % compact symbols for 1/2, etc.
\usepackage{microtype}      % microtypography
\usepackage{lipsum}
\usepackage{amssymb}
\usepackage{amsmath}
\usepackage{bm}
\usepackage[nocompress]{cite}
\usepackage{enumitem}
\usepackage{array}

\title{Quantifying the uncertainty of precision estimates for rule based text classifiers}

\author{
  James Nutaro \thanks{J. Nutaro and O. Ozmen are with the Computational Science and Engineering Division at Oak Ridge National Laboratory, Oak Ridge, TN, 32831 USA} \\
  \texttt{nutarojj@ornl.gov} \\
   \And
 Ozgur Ozmen \\
  \texttt{ozmeno@ornl.gov} \\
}

\begin{document}
\maketitle

\begin{abstract}
Rule based classifiers that use the presence and absence of key sub-strings to make classification decisions have a natural mechanism for quantifying the uncertainty of their precision. For a binary classifier, the key insight is to treat partitions of the sub-string set induced by the documents as Bernoulli random variables. The mean value of each random variable is an estimate of the classifier's precision when presented with a document inducing that partition. These means can be compared, using standard statistical tests, to a desired or expected classifier precision. A set of binary classifiers can be combined into a single, multi-label classifier by an application of the Dempster-Shafer theory of evidence. The utility of this approach is demonstrated with a benchmark problem.
\end{abstract}

% keywords can be removed

\section{Introduction}
\label{sect:intro}

The performance of machine learning applied to natural language processing tasks has seen significant improvement in recent years. Major advancements include word2vec \cite{mikolov2013efficient}, which generates embedding space based on co-occurrence of words in a fixed window; ELMo \cite{peters2018deep}, which uses a bidirectional recurrent neural network architecture; and BERT \cite{devlin2018bert} leveraging transformer encoders. There are also significant numbers of variants (see \cite{liu2019roberta,zhang2019dialogpt,howard2018universal}) of these architectures and other focusing on  benchmark text classification and sentiment analysis tasks.

Notwithstanding these improvements, state of the art methods to not promise consistent performance at a high level of precision. Even when high performance is achieved relative to a benchmark data set, strong evidence that a high level of performance will continue in practice cannot be offered. This is a critical barrier for natural language processing in domains, such as healthcare, where consistently good performance is essential.

A primary characteristic of machine learning is the very large number of free parameters that are fit to training data. Typically the number of parameters is on par with or exceeds the number of data points used for training. One consequence of so many free parameters is difficulty anticipating the performance of the network. A highly over parameterized model will fit the training data very closely, but this does not guarantee similar performance in practice.

Measures of excess parameters in a neural network have evolved for at least a decade. A method for applying the venerable minimal descriptive length criteria to neural networks was proposed by Hinton and van Camp \cite{Hinton93} and revised into a more computationally attractive form by Graves \cite{Graves11}. These and similar methods may point the way to smaller neural networks, but the number of parameters is unlikely become small enough that statistical tools can give insight into expected performance.

The observation that a useful neural network over fits its training data has led to fundamental questions concerning why we often see good performance when the network is presented with test data \cite{Belkin}. However, good results are not always the case, and the risks posed by a sudden failure to perform have also been a subject of considerable interest \cite{Harford,Hohle,DUNSON20184}. In spite of these questions and concerns, the future performance of state of the art methods in machine learning are guessed at via performance on testing data and hope that these tests are representative of performance in practice. 

The test and hope method has been very successful for a range of problems that may be described as low risk. In a low risk application, the consequences of an improper classification are small. If the test performance proves not to be representative of performance in practice, then the model is retracted and another put in its place. However, machine learning has been mostly absent from applications where consequences may be large and risks must be quantified.

Considerable effort has been expended to produce quantifiable guarantees for the performance of artificial neural networks. Where there has been success, it is with specific applications using highly specialized methods. For example, the novel training procedure developed for a neural network used as part of a flight control system \cite{Smith2010,Nguyen2010}, a novel weight updating technique to achieve pitch control in a submarine \cite{Song2010}, and a specialized training technique for modeling the blending of crude oils \cite{Li2010}. Though successful in their niche applications, the properties of these highly specialized networks cannot be generalized.

In a review of how artificial neural networks are used in the health care industry, Shahid et. al. \cite{Shahid} emphasizes that ``lack of transparency or interpretability of neural networks continues to be an important problem since health care providers are often unwilling to accept machine recommendations without clarity regarding the underlying rationale.'' In another survey, Xiao et. al. \cite{Xiao} remark that ``to bring deep models built from electronic health record data into real use, users often need to understand the mechanisms by which models operate. Such  a  level  of  model  transparency  is still  challenging  to achieve.'' Similar challenges are cited in many other reviews of artificial intelligence for health care; see, e.g., \cite{Jiang230,Finlayson1287,Challen231}.

Despite the tremendous activity directed at integrating machine learning into high reliability systems, this has not occurred to a substantial degree. It is conceivable that the very nature of the best machine learning methods, which leverage very large numbers of free parameters, may make them fundamentally unsuitable for high reliability applications. This possibility motivates the exploration of alternative techniques that offer a clear statistical basis for calculating risk.

We take a step towards statistically accessible classification by constructing a text classification scheme that has a natural basis for anticipating performance. Unsurprisingly, a key feature of our method is a small number of parameters, which prevents the classifier from performing as well on benchmarks as state of the art methods for machine learning. However, models created with the proposed method may be subject to rigorous statistical tests of performance without the use of distinct testing data. If testing data is available, then standard statistical tests can be used to determine if performance in test and training is the same. This capability for rigorous statistical analysis may permit the classifier to be used in applications that require quantifiable risk.

Our presentation begins with a description of how to build binary classifiers that use key word sets to distinguish documents, and we show how these may be interpreted as Bernoulli random variables for the purpose of statistical analysis. Having presented the binary classifiers, we show how they are merged into a multi-label classifier via combination in the Dempster-Shafer theory of evidence. Finally, we discuss possible extensions of the proposed model, including the use of word to vector models \cite{mikolov2013efficient} for picking appropriate key words.

\section{A statistically accessible classifier}

Genetic programming, in which simulated evolution discovers an effective classification rule, has been used to solve several text classification problems. The form of the evolved rule system is like those discussed by Pietramala et. al. \cite{Piet2008}, Hirsch et. al. \cite{Laur2005}, and Apt{\'e} et. al. \cite{Apte:1994}, which map a set $S$ of words to a classification decision. Genetic programming typically generates the set $S$ and a classification rule based on how a document partitions $S$ into words that are present and words that are not.

To concretely illustrate the correspondence between a classification rule and partitions of $S$, consider the documents in Table \ref{tab:example_documents}. The documents are strings labeled A or B. Suppose our classification rule labels a document B if it contains the strings bird or cat but not bright. This rule is a logical statement ``(bird or cat) and not bright''. The word set for this rule is
\begin{equation*}
S = \{ \text{bird}, \text{cat}, \text{bright} \} \text{ .}
\end{equation*}

There are eight partitions of this set. For the $k$th partition, let $I_k$ be the words appearing in the document and $O_k$ the words not in the document. Three partitions satisfy the rule, labelling the document B. These are
\begin{align*}
    I_1 & = \{ \text{bird}, \text{cat} \}, O_1=\{ \text{bright} \} \text{ ,} \\
    I_2 & = \{ \text{bird} \}, O_2=\{ \text{cat} , \text{bright} \} \text{ , and} \\
    I_3 & = \{ \text{cat} \}, O_3=\{ \text{bird} , \text{bright} \} \text{ .}
\end{align*}
The remaining five do not, and so a document inducing these partitions is labelled A. These partitions are
\begin{align*}
    I_4 & = \{ \text{cat}, \text{bird} , \text{bright} \}, O_4= \phi \text{ ,} \\
    I_5 & = \phi, O_5= \{\text{cat}, \text{bird} , \text{bright} \} \text{ ,} \\
    I_6 & = \{ \text{bright} \}, O_6= \{\text{cat}, \text{bird} \} \text{ ,} \\
    I_7 & = \{ \text{bright} , \text{cat} \}, O_7= \{ \text{bird} \} \text{ , and} \\
    I_8 & = \{ \text{bright} , \text{bird} \}, O_7= \{ \text{cat} \} \text{ .}
\end{align*}

For brevity we use $\pi_k$ to indicate the partition $\{ I_k, O_k \}$. For every document, there is a single $\pi_k$ that describes the disposition of key sub-strings in that document. These partitions and the consequent classification decision for our example documents are given on the right side of Table \ref{tab:example_documents}.
\begin{table}
\centering
\caption{Documents, induced partitions, and decisions of the classification rule}
\begin{tabular}{lcc}
    Document & $\pi_k$ & Classification \\
    The sun is bright & $\pi_6$ & A \\
    A bird sings brightly in the sunshine & $\pi_8$ & A \\
    The sun shines prettily & $\pi_5$ & A \\
    The stars twinkle brightly & $\pi_6$ & A \\
    The cat sings to the bird & $\pi_1$ & B \\
    The cat jumped over the moon & $\pi_3$ & B \\
    Two dozen birds doze in their cage & $\pi_2$ & B \\
    Two dozen cats watch two dozen birds & $\pi_1$ & B \\
    The sun, moon, and stars light up the sky & $\pi_5$ & A \\
    Carry the sun in a golden cup & $\pi_5$ & A \\
    The cat is out of the bag & $\pi_3$ & B \\
    That cat is a bright birder & $\pi_4$ & A \\
    The bright bird asked for a cracker & $\pi_7$ & A
\end{tabular}
\label{tab:example_documents}
\end{table}

\section{Partitions as random variables}
The $\pi_k$ are natural products of a classification rule. Moreover, they provide a statistical measure of its anticipated performance. Building on the example in Section \ref{sect:intro}, we may count how many times each $\pi_k$ is induced by a document labeled B and how many times it is induced by a document labeled A. 

Denote the count of A documents that induce partition $k$ by $a_k$ and the count of B documents by $b_k$. If we select a document at random and it induces partition $\pi_k$, then the probability of that document having label B is
\begin{equation*}
    p_k = \frac{b_k}{a_k+b_k} \text{ .}
\end{equation*}
This probability is the precision of the classification rule when it is used to detect documents labelled B that induce $\pi_k$. Indeed, each partition acts as a Bernoulli random variable that correctly indicates a document of type B with probability $p_k$.

More generally, when the $\pi_k$ are designed as binary classifiers, each document is labeled ``in class'' or ``out of class''. If the word set for the partition is chosen to identify documents that are ``in class'' then $p_k$ is the probability that a document inducing $\pi_k$ is ``in class'' and $1-p_k$ is the probability it is ``out of class''.

A key advantage of using the individual $\pi_k$ as classifiers is that standard tests of statistical significance characterize the uncertainty of $p_k$. A simple approach that works when $a_k+b_k$ is large and $p_k$ is away from zero or one is to approximate the Bernoulli random variable $\bm{\pi}_k$ with a normal random variable having mean $p_k$ and variance $p_k(1-p_k)$; see, e.g., \cite{Crow,Papoulis}.

The degrees of freedom in our estimate of $p_k$ is $a_k+b_k-1$. By knowing the degrees of freedom we may, for instance, use a one sided t-test to determine how likely it is that our classifier's precision fails to meet a minimum requirement $p$. The t-statistic for this test is
\begin{equation*}
    t = \frac{p_k-p}{
        \sqrt{
            \cfrac{p_k(1-p_k)}{a_k+b_k}
            }
        } \text{ .}
    \label{eqn:tstat}
\end{equation*}
We accept the classifier if $t$ is greater than a threshold that depends on the desired statistical significance and the degrees of freedom. This threshold can be calculated with a statistical program or looked up in a t-table (e.g., \cite{Crow,Papoulis}).

When applied to a classification problem, the t-test has two notable features. First, it supplies evidence with statistical significance $\alpha$ that documents inducing partition $\pi_k$ are labeled with precision $p$ or better. Hence, $\alpha$ and $p$ act as uncertainty metrics for our classifier; they indicate acceptable risk when applying the classification rule to a particular document. Second, we can decide if the classifier performs acceptably without resorting to a separate training corpus.

Explicit knowledge of the classification rules - in this case, the partitions of $S$ - is essential for any statistical test we might apply, and it is a non-trivial aspect of this type of classifier. Without this knowledge, the classifier is a black box for which $S$ and its partitions are unknown. The random variable describing this black box classifier as a whole comprises the possible partitions $\pi_1$, $...$, $\pi_n$ of $S$ and their relative frequencies of appearance (i.e., probabilities of appearance) $f_1$, $...$, $f_n$. It has the form
\begin{equation*}
    \mathbf{x} = \sum_{k=1}^n f_k \bm{\pi}_k \text{ .}
\end{equation*}

Now suppose we want to apply the one sided t-test to our estimate of this classifier's precision. There are $2n$ unknown parameters in $\mathbf{x}$ that influence this estimate. Specifically, the unknown $f_k$ and $p_k$ of the random variables $\bm{\pi}_k$. By using our training or testing data we obtain the sums $a=\sum a_k$ and $b=\sum b_k$, but the individual $a_k$ and $b_k$ are unknown. Our estimate of the precision of this classifier is $p = b/(a+b)$.

If we could look inside the box, then we would know the degrees of freedom in this estimate are $a+b-2n$. However, we cannot look into the box, we do not know $S$, and so we cannot know $n$. If we guess at $n$ and our guess is too small then we will be unjustifiably confident in our model. If our guess is too large, then we will be unjustifiably cautious. Regardless, the t-test is inconclusive.

Of course, not all possible partitions of a set $S$ will be induced by the available training documents. If one of these previously unseen partitions is later induced by a document that we wish to classify, that partition offers no information concerning the classification of the new document. Consequently, the partitions considered in the classification decisions are only those for which $a_k + b_k > 0$ after training the classifier.  

\section{Partitions as evidence}
A simple classification scheme that uses the $\pi_k$ individually would assign one or more labels to a document based on whether $p_k$ exceeds a given threshold. If exactly one label should be assigned to each document, then the relatively values of these probabilities should provide additional information about the likelihood of a particular label. The Dempster-Shafer theory provides a means of combining these probabilities when making a classification decision; a survey of its essential elements can be found in \cite{Kohlas1994}.

Within the Dempster-Shafer theory, we interpret the random variables $\bm{\pi}_k$ as sources of evidence, and this evidence is used to assign a label from the set of class labels $C = \{c_1, \dots , c_n \}$. Associated with the classes are word sets $S_1$, $\dots$ , $S_n$. We will consider the classification of a single document that induces partitions $\pi_1$, $\dots$ , $\pi_n$ with their associated probabilities $p_1$ , $\dots$ , $p_n$ of the document being ``in class''.

Let the set $C_k = \{ c_k \}$ so that its complement $\bar{C}_k$ contains every class label except $c_k$. For each partition $\pi_k$ we define a mass belief assignment $m_k$ such that for any $H \subset C$
\begin{equation*}
m_k(H) =
\begin{cases}
p_k & H = C_k \text{ ,} \\
1-p_k & H = \bar{C}_k \text{ , and} \\
0 & \text{otherwise.}
\end{cases}
\end{equation*}
Dempter's rule for combining evidence concerning the class $c_k$ is
\begin{equation}
    m(C_k) =\frac{\sum\limits_{H_1 \cap \dots \cap H_n = C_k} m_1(H_1) \dots m_n(H_n)}{1-\sum\limits_{H_1 \cap \dots \cap H_n = \phi}m_1(H_1) \dots m_n(H_n)}  \label{eqn:dempster}
\end{equation}
Each term within the numerator's sum is a product of $m_1$, $\dots$, $m_n$. The arguments $H_1$, $\dots$, $H_n$ in this product are subsets of $C$ such that their intersection is not empty. The denominator's sum is identical except that the arguments must have an empty intersection.

Looking first at the numerator, the single non-zero product involves $C_k \cap_{j \neq k} \bar{C}_j$. The corresponding term of the numerator is
\begin{equation*}
m_k(C_k) \prod_{j \neq k} m_j(\bar{C}_j) = p_k\prod_{j \neq k}(1-p_j) \text{ .}
\end{equation*}
For the denominator, the non-zero products are more numerous. The classes $c_1$, $\dots$ , $c_n$ have indices from the set $\{ 1,\dots, n \}$. For each $I \subset \{1, \dots , n\}$ with $|I| \neq 1$ the set $\cap_{j \in I} C_j \cap_{i \notin I} \bar{C}_i$ is empty and the product
\begin{equation*}
\prod_{j \in I} m_j(C_j) \prod_{i \notin I} m_i(\bar{C}_i) = \prod_{j \in I} p_j \prod_{i \notin I} (1-p_i)
\end{equation*}
may be greater than zero.
Substituting these into Eqn. \ref{eqn:dempster} produces
\begin{equation}
    m(C_k) = \frac{p_k \prod_{j \neq k}(1-p_j)}{1-\sum_{|I| \neq 1} \bigg ( \prod_{j \in I} p_j \prod_{i \notin I} (1-p_i) \bigg) } \text{ .}
    \label{eqn:hard}
\end{equation}
The belief (support) and plausibility of the hypothesis $C_k$ are upper and lower measures of how the individual $m_j$ collectively support a particular conclusion. In our specific case,
\begin{align*}
    & bel(C_k) = \sum\limits_{H \subset C_k} m(H) = m(C_k) \text{ and} \\
    & pl(C_k) = 1-bel(\bar{C}_k) = 1- \sum\limits_{j \neq k} m(C_j) \text{ .}
\end{align*}

These belief and plausibility measures have natural interpretations as upper and lower bounds for the likelihood that a document of class $c_k$ generates the particular partitions $\pi_1$, $\dots$, $\pi_n$. If the random variables $\bm{\pi}_1$, $\dots$, $\bm{\pi}_n$ are independent, then the product $p_k \prod_{j \neq k} (1-p_j)$ in the numerator of $bel$ is likelihood that the document is of class $c_k$ and not any other. However, we may discount the probability mass assigned to events that are definitely impossible.

One of these impossible events is that the document belongs to no class, which is to say it belongs to $\bar{C}_k$ for every $k$; that is, when $I$ is empty. The others are a document belonging to more than one class, and this is captured by $|I| > 1$. These are the subtracted quantities in the denominator of $m(C_k)$.

The plausibility of $C_k$ is what is left after we subtract the belief assigned to all other possibilities. Those alternatives with positive mass are the $C_j$, $j \neq k$. Since the sum of all possibilities is one, $pl(C_k)$ must be one minus the sum of the alternatives.

The $bel$ function is pessimistic because it considers only $m(C_k)$, which is the small possibility of seeing what was actually observed assuming perfect independence of the partitions. The $pl$ function is optimistic because it considers everything but $m(C_k)$, and so implicitly assumes perfect independence of the partitions other than $\pi_k$, but not necessarily the independence of that partition. The truth is somewhere in between.

It is infeasible to compute the denominator in Eqn. \ref{eqn:hard} directly when $n$ is large. Fortunately, it is not necessary to do so. For any $H \neq C_i$, $i = 1, \dots, n$, $m(H) = 0$. A property of $m$ is $\sum_{H \subset C} m(H) = 1$ \cite{Kohlas1994}. Moreover, the denominators of the $m(C_i)$ are identical. Therefore
\begin{align*}
     \sum_{i = 1}^n m(C_i) & = 1  \text{, and so} \\
     1-\sum_{|I| \neq 1} \bigg ( \prod_{j \in I} p_j \prod_{i \notin I} (1-p_i) \bigg) & = \sum_{i=1}^n \bigg ( p_i \prod_{j \neq i}(1-p_j) \bigg ) \text{ .}
\end{align*}
This gives us the computationally feasible expression
\begin{equation}
m(C_k) = \frac{p_k \prod_{j \neq k}(1-p_j)}{\sum_{i=1}^n \bigg ( p_i \prod_{j \neq i}(1-p_j) \bigg ) } \text{ .}
\label{eqn:feasible}
\end{equation}

The above observation has a curious effect on belief and plausibility. Because the $m(C_k)$ sum to one, it is necessary that $pl(C_k) = bel(C_k)$. Hence, if $pl$ and $bel$ are boundaries for some probability, then Eqn. \ref{eqn:feasible} gives us the best possible estimate of that probability. Furthermore, if we desire a classifier with precision $q$ and restrict our classification decisions to documents that induce $m(C_k) > q$ then this gated classifier will exhibit at least the desired precision at the cost of a reduction in recall.

\section{An example}
\label{sect:example}
We demonstrate these statistical concepts by evolving classifiers for the Reuters-21578 benchmark after removing documents with multiple labels. Two specific demonstrations are given. In the first, we use a two sided t-test to examine the difference in precision of the classification rule for the three most populous classes when that classifier is applied to training and testing data. For this purpose we use the typical ModeApte split into testing and training documents \cite{Apte:1994}. The most populous classes were selected to ensure that $\bm{\pi}_k$ is approximately normal. In the second demonstration, we compare the precision and recall achieved when classification decisions are gated by a threshold value on $p_k$ and plausibility. 

A simple type of evolutionary search is used to construct a word set $S$ for each class given $d$ documents that are in class and some number of out of class documents. The objective function attempts to maximize the recall and precision when counting just documents that do not induce $I_k = \phi$.

Let $t(S)$ be the count of documents that are in the class of interest and induce a partition with $I_k \neq \phi$, and let $f(S)$ be the count of documents not in the class of interest that induce $I_k \neq \phi$. The goal is to maximize $F(S)$ where $F$ is defined by
\begin{align*}
    F(S) & = \frac{2P(S)R(S)}{P(S)+R(S)} \\
    P(S) & = \frac{t(S)}{t(S)+f(S)} \\
    R(S) & = \frac{t(S)}{d} \text{ .}
\end{align*}
When two solutions have the same fitness we prefer the one with fewer sub-sequences and then the solution with longer average sub-sequence length.

Prior to training, we convert all letters to lower case, replace anything not a-z with white space, and then compress sequences of white space to a single white space. We limit $S$ to two sub-strings of no more than fifteen characters. The training algorithm is as follows.
\begin{enumerate}
    \item Construct $S$ at random from sub-strings found in the documents.
    \item Create a new solution $S'$ by adding, removing, or changing a sub-string in $S$. Sub-strings in $S'$ are selected with equal chance from documents in class and out of class. \label{step:build}
    \item If $F(S') > F(S)$ then replace $S$ with $S'$.
    \item Go to Step \ref{step:build}.
\end{enumerate}
Table \ref{tab:reuters_S} shows the contents of $S$ for the three most populous classes. 

\begin{table}
    \centering
    \caption{Sub-strings for the Reuters classifiers}
    \begin{tabular}{l|l}
        class & words \\ \hline
        earn & `` cts'', `` net ''  \\
        acq & `` stake'', `` acq''  \\
        crude & `` crude'', ``arrel''
    \end{tabular}
    \label{tab:reuters_S}
\end{table}

To compare test and training results we use the two sided t-test. If $d'_k$ and $d_k$ are the counts of documents in the class for testing and training respectively, and likewise for the counts $b'_k$ and $b_k$ of documents out of class, the t statistic is
\begin{align*}
    t & = \frac{p_k-p'_k}{\sqrt{\hat{p}_k(1-\hat{p}_k)\bigg ( \cfrac{1}{n'_k}+\cfrac{1}{n_k}} \bigg )} \\
    n'_k & = d'_k+b'_k \\
    p'_k & = d'_k/n'_k \\
    n_k & = d_k+b_k \\
    p_k & = d_k/n_k \\
    \hat{p}_k & = (d'_k+d_k)/(n'_k+n_k)
\end{align*}
and we compute critical values $t_{\text{crit}}$ at the $\alpha=0.05$ significance level with degrees of freedom $f=\min\{n'_k,n_k\}-1$. If $|t| > t_{\text{crit}}$ then we reject the hypothesis that $p'_k=p_k$ and accept it otherwise \cite{Papoulis,Crow}.

Training and testing precision and the results of the two-sided t-test are given in Table \ref{tab:reuters_P}. The counts used to calculate these statistics are in Table \ref{tab:reuters_D}. The column headed ``partition'' contains bit strings with a $1$ if the word is present and zero if not; the bits are ordered just as the words are in Table \ref{tab:reuters_S}.

\begin{table}[!ht]
    \centering
    \caption{Test and training results for $p_k$}
    \begin{tabular}{llllllll}
        class & partition & $p$ & $p'$ & $f$ & $t_{\text{crit}}$ & $t$ & $p \neq p'$ \\ \hline
    earn & 00 & 0.108 & 0.0510 & 1451 & 1.96 & 6.38 & yes \\
        & 01 & 0.748 & 0.743 & 182 & 1.96 & 0.133 & no \\
        & 10 & 0.893 & 0.591 & 175 & 1.96 & 9.83 & yes \\
        & 11 & 0.995 & 0.996 & 771 & 1.96 & -0.214 & no \\ \hline
    acq & 00 & 0.121 & 0.130 & 2088 & 1.96 & -1.05 & no \\
        & 01 & 0.812 & 0.842 & 347 & 1.96 & -1.21 & no \\
        & 10 & 0.817 & 0.855 & 82 & 1.99 & -0.786 & no \\
        & 11 & 0.941 & 0.968 & 62 & 2.00 & -0.813 & no \\ \hline
    crude & 00 & 0.0111 & 0.0101 & 2473 & 1.96 & 0.388 & no \\
        & 01 & 0.731 & 0.625 & 23 & 2.07 & 0.994 & no \\
        & 10 & 0.522 & 0.903 & 30 & 2.04 & -3.68 & yes \\
        & 11 & 0.909 & 0.981 & 53 & 2.01 & -1.73 & no
    \end{tabular}
    \label{tab:reuters_P}
\end{table}

\begin{table}[!ht]
    \centering
    \caption{Counts for each partition}
    \begin{tabular}{ll|cc|cc}
    class & partition & $d_k$ & $b_k$ & $d'_k$ & $b'_k$ \\ \hline
    earn & 00 & 424 & 3514 & 74 & 1378 \\
        & 01 & 398 & 134 & 136 & 47 \\
        & 10 & 696 & 83 & 104 & 72 \\
        & 11 & 1322 & 6 & 769 & 3 \\ \hline
    acq & 00 & 658 & 4787 & 271 & 1818 \\
        & 01 & 622 & 144 & 293 & 55 \\
        & 10 & 188 & 42 & 71 & 12 \\
        & 11 & 128 & 8 & 61 & 2 \\ \hline
    crude & 00 & 70 & 6261 & 25 & 2449 \\
        & 01 & 57 & 21 & 15 & 9 \\
        & 10 & 36 & 33 & 28 & 3\\
        & 11 & 90 & 9 & 53 & 1
    \end{tabular}
    \label{tab:reuters_D}
\end{table}

The hypothesis $p=p'$ is rejected in three cases: the category earn with none of the selected sub-strings; earn with `` cts'' in and `` net '' out; and crude with `` crude'' in and ``arrel'' out. All three cases almost certainly reflect genuine differences between the testing and training data as they would also be rejected at much higher confidence levels.

One explanation for the rejections is that the variety of possible documents is not adequately represented in the testing documents, training documents, or both \cite{Ashby}. If so, a suitably informed subject matter expert might resolve the question of rejecting the classifier or attributing the test result to a type II error. Having the option of requesting expert opinion concerning this issue is an important advantage of knowing the classification rule.

Regardless, it is clear that the model does not suffer from over-fitting. Notably, the hypothesis $p=p'$ is rejected once with $p' > p$ and twice with $p' < p$. That is, the testing results may be better or worse than the training results when our hypothesis is rejected.

Our measures of expected precision in the form of $p_k$ and plausibility allow us to be selective when classifying documents and thereby achieve a desired level of precision at the cost of reducing recall. Given a desired precision $q$, we assign a label $c_k$ only if $p_k > q$ or if $pl(C_k) > q$. If our estimates of these quantities are good, then we will achieve the desired level of performance. 

Table \ref{tab:my_label} illustrates the effect of this gated classification decision when $q = 0.9$. Two scores are reported. The first case is in the row labeled $p$. Here we assign the document to the class with the highest $p_k$ if that $p_k > q$. The second case assigns the label with the highest $pl(C_k)$ if that $pl(C_k) > q$. The number preceding the slash is for the training data and the number following is for the testing data. 

\begin{table}[!ht]
    \centering
    \caption{Precision and recall}
    \label{tab:my_label}
    \begin{tabular}{c|c|c|c}
        class & measure & Precision & Recall \\
        earn & $p > 0$ &  0.959 / 0.939 & 0.827 / 0.910 \\
             & $p > 0.9$ & 0.995 / 0.996 & 0.464 / 0.704 \\
        acq  & $p > 0$ & 0.601 / 0.627 & 0.962 / 0.948 \\
             & $p > 0.9$ & 0.955 / 0.968 & 0.0796 / 0.0876 \\
        crude & $p > 0$ & 0.903 / 0.937 & 0.660 / 0.612 \\
              & $p > 0.9$ & 0.968 / 0.981 & 0.356 / 0.430 \\ \hline
        earn & $pl > 0$ &  0.959 / 0.939 & 0.827 / 0.916 \\
             & $pl > 0.9$ & 0.978 / 0.971 & 0.814 / 0.908 \\
        acq  & $pl > 0$ & 0.602 / 0.627 & 0.962 / 0.948 \\
             & $pl > 0.9$ & 0.943 / 0.968 & 0.530 / 0.562 \\
        crude & $pl > 0$ & 0.902 / 0.937 & 0.656 / 0.612 \\
              & $pl > 0.9$ & 1 / 1 & 0.233 / 0.190
    \end{tabular}
\end{table}

The performance in these two cases is essentially the same when we attempt to classify every document; this is the $> 0$ case in the table. The metrics become more distinct when we gate the classification decision. The observed precision when applying a threshold to $p_k$ is close to the precision anticipated in Table \ref{tab:reuters_P}. The primary effect of the threshold is to discard documents that induce partitions with low precision. A small deviation from the anticipated performance occurs as a result of comparing percentages across all classes and choosing the most probably.

The main effect of using plausibility is to boost recall. This occurs for the earn and acq classes. The much smaller crude class sees a drop in recall percentage, but an examination of Table \ref{tab:reuters_D} suggests the reduction in number of documents is small.

For the large acq and earn classes, the boost is expected because the plausibility is an upper bound on the expected precision of the classifier. Indeed, because $pl=bel$ we also expect it to be the lower bound. Notably, the observed precision is above the chosen threshold and there is no indication of over fitting. Where testing and training precision are not close, either may be the more precise.

\section{Conclusion}
We have shown how a classifier that uses a set of sub-strings and a rule acting on the presence and absence of those sub-strings has a natural, statistical mechanism for quantifying uncertainty in the classification of a document. A natural use of the one and two sided t-tests would be in a two step process. In the first step, separate training and testing documents are used to build the classifier and check, via the two sided t-test, that it performs identically in testing and training. When this hypothesis is rejected, a subject matter expert reviews the classification criteria to decide if the rejection is warranted.

In the second step, a classifier is built using all of the documents. This classifier is compared to the classifier from the first step. If new classification rules appear, then these may be accepted, rejected for lack of testing, or reviewed by a subject matter expert. The one sided t-test is used to check that the new classifier has acceptable accuracy. When put into use, the classifier's continuing suitability may be monitored by tests for statistical agreement between the expected precision and the precision observed in small sets of documents that are manually classified for quality control purposes.

We have focused on the simple case of documents with a single label, but this is not a requirement of the Dempster-Shafer theory. If multiple labels are allowed, it would be necessary to redefine the $m_k$ such that they assign belief mass to multi-label sets; e.g., $m_k(\{c_1,c_2\}) > 0$ would be necessary if both $c_1$ and $c_2$ could be assigned to a single document. Given a suitable definition of $m_k$, it would be necessary to reconsider which subsets of $C$ in Eqn. \ref{eqn:dempster} have positive mass. This could quickly become very computationally challenging as the number of possible subsets grows.

A similar, unexplored possibility is exploiting the distribution of document classes associated with each partition. The distinction between in class and out of class is rather coarse. In fact, each partition defines a discrete probability density function over all of the document classes. To take advantage of this information would require a new definition of $m_k$ that assigns a distinct belief mass to each class which induces $\pi_k$. As before, this would require a reexamination of Eqn. \ref{eqn:dempster} and could lead to a computationally challenging belief function.

Finally, we remark on the possible use of word to vector models to select key sub strings. To begin, note that is not necessary to identify a single set $S$. We could just as easily consider word sets $S_1$, $S_2$, $...$, $S_n$. A document would induce partitions of each set, and these partitions can be treated as evidence, just as before.

For instance, suppose we wish to identify documents that discuss Czech currency. We would use $\{$ czech , currency $\}$ as one keyword set. The other sets would be constructed from words with vectors near the vector sum $vec(\text{czeck})+vec(\text{currency})$. Using the results from Table 5 in \cite{mikolov2013efficient} this would yield $\{$koruna$\}$, $\{$czeck crown$\}$, $\{$polish zolty$\}$, and $\{$ctk$\}$; or we might choose some other arrangement of the nearby words and phrases into keyword sets. If this approach is found to be effective, it could offer a much more computationally efficient key word search than was used in Sect. \ref{sect:example}.

A key feature of the proposed classification method is knowledge of the mutually exclusive feature sets that lead to a classification decision. Recent work on attention in neural networks, such as the steps towards interpreting network decisions described by Brown et. al. \cite{Brown:2018}, may someday offer the possibility of clearly identifying which features are chosen during network training. If so, each combination of features would be associated with a distribution of classification outcomes, just as the partitions of $S$ can be treated as Bernoulli random variables.

Given mutually exclusive feature sets and their distributions of outcomes, it would be natural to extend the statistical methods described here to this new application. The main obstacle appears to be ensuring mutual exclusivity or, absent that, a reliable measure of degrees of freedom in the data used to estimate the probability distributions.

\section*{Acknowledgements}
This manuscript has been authored by UT-Battelle, LLC under Contract No. DE-AC05-00OR22725 with the U.S.~Department of Energy. The United States Government retains and the publisher, by accepting the article for publication, acknowledges that the United States Government retains a non-exclusive, paid-up, irrevocable, world-wide license to publish or reproduce the published form of the manuscript, or allow others to do so, for United States Government purposes. The Department of Energy will provide public access to these results of federally sponsored research in accordance with the DOE Public Access Plan (\url{http://energy.gov/downloads/doe-public-access-plan}).

This work has been supported in part by the Joint Design of Advanced Computing Solutions for Cancer (JDACS4C) program established by the U.S. Department of Energy (DOE) and the National Cancer Institute (NCI) of the National Institutes of Health. This work was performed under the auspices of the U.S. Department of Energy under Contract DE-AC05-00OR22725.

\bibliographystyle{unsrt}  
\bibliography{refs}  %%% Remove comment to use the external .bib file (using bibtex).

\begin{thebibliography}{10}

\bibitem{mikolov2013efficient}
Tomas Mikolov, Kai Chen, Greg Corrado, and Jeffrey Dean.
\newblock Efficient estimation of word representations in vector space, 2013.

\bibitem{peters2018deep}
Matthew~E Peters, Mark Neumann, Mohit Iyyer, Matt Gardner, Christopher Clark,
  Kenton Lee, and Luke Zettlemoyer.
\newblock Deep contextualized word representations, 2018.

\bibitem{devlin2018bert}
Jacob Devlin, Ming-Wei Chang, Kenton Lee, and Kristina Toutanova.
\newblock Bert: Pre-training of deep bidirectional transformers for language
  understanding, 2018.

\bibitem{liu2019roberta}
Yinhan Liu, Myle Ott, Naman Goyal, Jingfei Du, Mandar Joshi, Danqi Chen, Omer
  Levy, Mike Lewis, Luke Zettlemoyer, and Veselin Stoyanov.
\newblock Roberta: A robustly optimized bert pretraining approach, 2019.

\bibitem{zhang2019dialogpt}
Yizhe Zhang, Siqi Sun, Michel Galley, Yen-Chun Chen, Chris Brockett, Xiang Gao,
  Jianfeng Gao, Jingjing Liu, and Bill Dolan.
\newblock Dialogpt: Large-scale generative pre-training for conversational
  response generation, 2019.

\bibitem{howard2018universal}
Jeremy Howard and Sebastian Ruder.
\newblock Universal language model fine-tuning for text classification, 2018.

\bibitem{Hinton93}
Geoffrey~E. Hinton and Drew van Camp.
\newblock Keeping the neural networks simple by minimizing the description
  length of the weights.
\newblock In {\em Proceedings of the Sixth Annual Conference on Computational
  Learning Theory}, COLT ’93, page 5–13, New York, NY, USA, 1993.
  Association for Computing Machinery.

\bibitem{Graves11}
Alex Graves.
\newblock Practical variational inference for neural networks.
\newblock In J.~Shawe-Taylor, R.~S. Zemel, P.~L. Bartlett, F.~Pereira, and
  K.~Q. Weinberger, editors, {\em Advances in Neural Information Processing
  Systems 24}, pages 2348--2356. Curran Associates, Inc., 2011.

\bibitem{Belkin}
Mikhail Belkin, Siyuan Ma, and Soumik Mandal.
\newblock To understand deep learning we need to understand kernel learning.
\newblock In {\em Proceedings of the 35th International Conference on Machine
  Learning}, volume~80, pages 531--540, Stockholm, Sweden, 2018. PMLR.

\bibitem{Harford}
Tim Harford.
\newblock Big data: A big mistake?
\newblock {\em Significance}, 11(5):14--19, 2014.

\bibitem{Hohle}
Michael H{\"o}hle.
\newblock A statistician's perspective on digital epidemiology.
\newblock {\em Life Sciences, Society and Policy}, 13, 2017.

\bibitem{DUNSON20184}
David~B. Dunson.
\newblock Statistics in the big data era: Failures of the machine.
\newblock {\em Statistics \& Probability Letters}, 136:4 -- 9, 2018.
\newblock The role of Statistics in the era of big data.

\bibitem{Smith2010}
Tim Smith, Jim Barhorst, and James~M. Urnes.
\newblock {\em Design and Flight Test of an Intelligent Flight Control System},
  pages 57--76.
\newblock Springer Berlin Heidelberg, Berlin, Heidelberg, 2010.

\bibitem{Nguyen2010}
Nhan~T. Nguyen and Stephen~A. Jacklin.
\newblock {\em Stability, Convergence, and Verification and Validation
  Challenges of Neural Net Adaptive Flight Control}, pages 77--110.
\newblock Springer Berlin Heidelberg, Berlin, Heidelberg, 2010.

\bibitem{Song2010}
Y.~D. Song, Liguo Weng, and Medorian~D. Gheorghiu.
\newblock {\em Pitch-Depth Control of Submarine Operating in Shallow Water via
  Neuro-adaptive Approach}, pages 165--178.
\newblock Springer Berlin Heidelberg, Berlin, Heidelberg, 2010.

\bibitem{Li2010}
Xiaoou Li and Wen Yu.
\newblock {\em Modeling of Crude Oil Blending via Discrete-Time Neural
  Networks}, pages 205--220.
\newblock Springer Berlin Heidelberg, Berlin, Heidelberg, 2010.

\bibitem{Shahid}
Nida Shahid, Tim Rappon, and Whitney Berta.
\newblock Applications of artificial neural networks in health care
  organizational decision-making: A scoping review.
\newblock {\em PLOS ONE}, 14(2):1--22, 02 2019.

\bibitem{Xiao}
Cao Xiao, Edward Choi, and Jimeng Sun.
\newblock {Opportunities and challenges in developing deep learning models
  using electronic health records data: a systematic review}.
\newblock {\em Journal of the American Medical Informatics Association},
  25(10):1419--1428, 06 2018.

\bibitem{Jiang230}
Fei Jiang, Yong Jiang, Hui Zhi, Yi~Dong, Hao Li, Sufeng Ma, Yilong Wang, Qiang
  Dong, Haipeng Shen, and Yongjun Wang.
\newblock Artificial intelligence in healthcare: past, present and future.
\newblock {\em Stroke and Vascular Neurology}, 2(4):230--243, 2017.

\bibitem{Finlayson1287}
Samuel~G. Finlayson, John~D. Bowers, Joichi Ito, Jonathan~L. Zittrain,
  Andrew~L. Beam, and Isaac~S. Kohane.
\newblock Adversarial attacks on medical machine learning.
\newblock {\em Science}, 363(6433):1287--1289, 2019.

\bibitem{Challen231}
Robert Challen, Joshua Denny, Martin Pitt, Luke Gompels, Tom Edwards, and
  Krasimira Tsaneva-Atanasova.
\newblock Artificial intelligence, bias and clinical safety.
\newblock {\em BMJ Quality \& Safety}, 28(3):231--237, 2019.

\bibitem{Piet2008}
Adriana Pietramala, Veronica~L. Policicchio, Pasquale Rullo, and Inderbir
  Sidhu.
\newblock A genetic algorithm for text classification rule induction.
\newblock In Walter Daelemans, Bart Goethals, and Katharina Morik, editors,
  {\em Machine Learning and Knowledge Discovery in Databases}, pages 188--203,
  Berlin, Heidelberg, 2008. Springer Berlin Heidelberg.

\bibitem{Laur2005}
Laurence Hirsch, Masoud Saeedi, and Robin Hirsch.
\newblock Evolving rules for document classification.
\newblock In Maarten Keijzer, Andrea Tettamanzi, Pierre Collet, Jano van
  Hemert, and Marco Tomassini, editors, {\em Genetic Programming}, pages
  85--95, Berlin, Heidelberg, 2005. Springer Berlin Heidelberg.

\bibitem{Apte:1994}
Chidanand Apt{\'e}, Fred Damerau, and Sholom~M. Weiss.
\newblock Automated learning of decision rules for text categorization.
\newblock {\em ACM Transactions on Information Systems}, 12(3):233--251, July
  1994.

\bibitem{Crow}
Edwin~L. Crow, Frances~A. Davis, and Margaret~W. Maxfield.
\newblock {\em Statistics manual}.
\newblock Dover publications, inc., New York, NY, USA, 1960.

\bibitem{Papoulis}
Athanasios Papoulis.
\newblock {\em Probability, random variables, and stochastic processes, 3rd
  edition}.
\newblock McGraw-Hill, Inc., New York, NY, USA, 1991.

\bibitem{Kohlas1994}
J{\"u}rg Kohlas and Paul-Andr{\'e} Monney.
\newblock Theory of evidence --- a survey of its mathematical foundations,
  applications and computational aspects.
\newblock {\em Zeitschrift f{\"u}r Operations Research}, 39(1):35--68, Feb
  1994.

\bibitem{Ashby}
W.R. Ashby.
\newblock Requisite variety and its implications for the control of complex
  systems.
\newblock {\em Cybernetica}, 1(2):83--99, 1958.

\bibitem{Brown:2018}
Andy Brown, Aaron Tuor, Brian Hutchinson, and Nicole Nichols.
\newblock Recurrent neural network attention mechanisms for interpretable
  system log anomaly detection.
\newblock In {\em Proceedings of the First Workshop on Machine Learning for
  Computing Systems}, MLCS'18, pages 1:1--1:8, New York, NY, USA, 2018. ACM.

\end{thebibliography}

\end{document}